# Novel Meta-Heuristic Model for Discrimination between Iron Deficiency Anemia and B-Thalassemia with CBC Indices Based on Dynamic Harmony Search (DHS)

Sultan Noman Qasem[1,2], and Amir Mosavi[3*]

[1]Computer Science Department, College of Computer and information Sciences, Al Imam Mohammad Ibn Saud Islamic University (IMSIU), Riyadh, Saudi Arabia
[2]Computer Science Department, Faculty of Applied Science, Taiz University, Taiz, Yemen
[3]Kalman Kando Faculty of Electrical Engineering, Obuda University, 1034 Budapest, Hungary

Corresponding author: amir.mosavi@kvk.uni-obuda.hu

**ABSTRACT** In recent decades, attention has been directed at anemia classification for various medical purposes, such as thalassemia screening and predicting iron deficiency anemia (IDA). In this study, a new method has been successfully tested for discrimination between IDA and β-thalassemia trait (β-TT). The method is based on a Dynamic Harmony Search (DHS). Complete blood count (CBC), a fast and inexpensive laboratory test, is used as the input of the system. Other models, such as a genetic programming method called structured representation on genetic algorithm in non-linier function fitting (STROGANOFF), an artificial neural network (ANN), an adaptive neuro-fuzzy inference system (ANFIS), a support vector machine (SVM), k-nearest neighbor (KNN), and certain traditional methods, are compared with the proposed method.

**INDEX TERMS** Anemia classification, dynamic harmony search, iron deficiency anemia, thalassemia trait,

## I. INTRODUCTION

Iron deficiency anemia (IDA) and β-thalassemia trait (β-TT) are the most common types of anemia. They have similar effects on red blood cell indices such as red blood cell count (RBC), hemoglobin concentration (Hb), mean corpuscular volume (MCV), mean corpuscular hemoglobin (MCH), mean corpuscular hemoglobin concentration (MCHC), red blood cell distribution width (RDW), and hematocrit (HCT). Thalassemia disease is a genetic trait that directly impacts on reducing the life span of red blood cells through regulating the formation of hemoglobin (Hb) as a fundamental component of red blood cells [1]. In contrast, IDA is not inherited and can generally be treated by dietary changes and iron supplements. Complete blood count (CBC) is the primary clinical test for diagnosis of anemia. If CBC indices cannot discriminate between IDA and β-TT, complementary tests, such as hemoglobin electrophoresis or genetic tests, are needed. Complementary tests are time consuming, expensive and not available in all laboratories and hospitals. As a result, scientists have conducted many studies on the discrimination between IDA and β-TT anemia classification using the indices of the CBC test [2].

Historically, mathematical formulas have been used to solve the problem [3], [4], [13]–[16], [5]–[12]. Recently, there has been increasing interest in the use of artificially intelligent automated medical diagnostic systems [17]–[22]. There exists engineering applications for anemia classification including image analysis [23], statistical analysis [24], and clustering techniques [25], rule-based expert systems [26]–[28], hybrid neural rule-based method [29]. One of the first efforts to diagnosis anemia using an ANN was performed by Brindorf et al. in 1996 in which a hybrid expert system of rule-based and ANN has been developed. They reported 96.5% precision in classifying microcytic anemia and the anemia of chronic disease. In 2002, Nikolaev and de-Menezes introduced an image recognition approach for classification of anemia [30]. Amendolia et al. in 2003 published a comparative study of K-nearest neighbor, support vector machine (SVM) and multi-layer perceptron (MLP) methods for thalassemia screening [31], [32]. The results revealed that MLP slightly improved the classification sensitivity (92%) in comparison with SVM (83%) while the classification accuracies were equal to 95%. Another comparison has been performed by Azarkhish et al. in 2012 [33]. They compare three different methods, an adaptive neuro-fuzzy inference system (ANFIS), an ANN and a logistic regression model in diagnosing iron deficiency anemia (IDA), and reported that the ANN is superior to the others. All of these methods have been successfully tested on various types of anemia. Moreover, to narrow the domain of anemia specifically on β-TT and IDA, an alternative diagnostic tool was compulsory. This study proposes such a tool. It compares the existing methods for discriminating β-TT and IDA, including traditional methods and the newer, AI-based diagnostic systems, with a novel meta-heuristic model. The main goal of this study is to reduce the diagnosis time and cost for β-TT and IDA subjects by increasing their discrimination precision through the analysis of CBC indices.

This paper is organized as follows. Section 2 describes anemia and its effects on blood indices. Section 3 introduces the harmony search (HS) algorithm. In section 4, materials and methods are presented. Implementation details, experimental tests, and the results of the proposed method are discussed in section 5. Further tests and comparisons with other methods and models are presented in section 6. Finally, section 7 presents the study's conclusions.

## II. EFFECTS OF ANEMIA ON CBC INDICES

There are several different, task-specific blood cell types, such as white blood cells (WBCs), red blood cells (RBCs), and platelets. A red blood cell transmits oxygen to all cells of the body and receives the $CO_2$ excreted by them. These tasks are carried out by hemoglobin. Each red blood cell contains approximately 300 million molecules of hemoglobin. Hemoglobin consists of four polypeptide chains (globin) and heme. Hence, a change in the structure of globin affects the structure and functionality of the red blood cells. Each molecule of hemoglobin has four heme molecules. A heme molecule is a cofactor consisting of an $Fe^{2+}$ (ferrous) ion in the center of a heterocyclic organic ring called porphyrin. The iron atom can bind an oxygen molecule through ion-induced dipole forces. In 1825, J.F. Engelhard discovered that the ratio of Fe to protein is identical in the hemoglobin of several species [34]. Iron deficiency anemia (IDA) leads to a reduction in the quantity and the quality of hemoglobin.

There are four types of globin chains: alpha (α), beta (β), gamma (γ), and delta (δ). Based on the type of globin chains in its structure, the hemoglobin belongs to one of three major types, HbA (α2β2), HbA2 (α2δ2), or HbF (α2γ2). HbF, or fetal hemoglobin, is the main component of fetal red blood cells, which gradually diminishes in proportion to HbA after birth. The proportions of HbA, HbA2, and HbF in a mature, healthy person's blood are approximately 97%, 2-3%, and less than 1%, respectively.

Based on the proportions of the hemoglobin types in a person, HBA (α2β2) is a major component of red blood cells. As a result, the most vital globin types are α-globin and β-globin. The first contains 141 amino acids regulated by genes on chromosome 16 and the second contains 161 amino acids regulated by genes on chromosome 11. Thalassemia results from defects on the genes that regulate the formation of globin chains. The type of thalassemia and the severity of the disease depend on the types and the number of defective genes, respectively. A person must have at least two defective genes of the same type to have the disease. If only one defective gene or two different defective genes are present, he has thalassemia trait. Persons with thalassemia trait do not have the disease but inherit genes that cause the disease. The type of defective genes determines whether the subject has α-thalassemia or β-thalassemia. Generally, β-thalassemia is more common than α-thalassemia.

### A. IRON DEFICIENCY ANEMIA (IDA)

As previously discussed, IDA leads to a reduction in heme. It is clear that a reduction in heme affects the concentration of hemoglobin in blood. Therefore, in a person with IDA, the blood parameters dependent on the hemoglobin concentration are affected. Some of these parameters, which are generally used as diagnosis indices, are: hemoglobin concentration (Hb), which denotes the quantity of hemoglobin in blood, red blood cell count (RBC), mean corpuscular volume (MCV), mean corpuscular hemoglobin (MCH), mean corpuscular hemoglobin concentration (MCHC), red blood cell distribution width (RDW), and hematocrit (HCT).

### B. CHARACTERISTIC OF THALASSEMIA

In α-thalassemia or β-thalassemia, regardless of whether the subject has the disease or the trait, hemoglobin concentration becomes lower than normal. This reduction is a result of a decreased concentration of α-globin (α-thalassemia) or β-globin (β-thalassemia) chains. When the concentration of α or β-globin reduces, the quantity and quality of hemoglobin are naturally lower than normal. As a result, similar to an IDA subject, someone with thalassemia disease or trait has reduced Hb, MCV, MCH, MCHC, and HCT values. For thalassemia disease, diagnosis is not difficult because the indices are too low. But in thalassemia trait, the indices are only a little below the normal range, leading to difficulty discriminating between IDA and thalassemia trait.

One of the most important differences between IDA and β-TT is the red blood cell count (RBC). Generally, to compensate for the lower Hb with β-TT, the RBC increases, while the RBC index is lower than normal for IDA. In summary, we can conclude that IDA and β-TT both lead to a reduced concentration of hemoglobin, but the reduction for β-TT is greater than for IDA. Despite these differing amounts of reduction, in some circumstances it is too difficult to discriminate between these types of anemia.

### III. HARMONY SEARCH ALGORITHM (HS)

Mathematical based optimization algorithms exploit the benefit of substantial gradient data. In addition, there exist the wide range of programming approaches including Linear and non-linear programming to solve optimization problems [35]. The initial population should have been selected carefully for optimal convergence of the optimization algorithm. In contrast, Non-gradient meta-heuristic methods are other popular solutions which does not have inherent difficulties of gradient-based methods [35]–[43]. Generally, meta-heuristic algorithms comes from nature and are guided random search algorithm [43]. Harmony search (HS) is inspired from jazz musicians and has variety of applications in engineering introduced by Geem Z-W. in 2001 [35] and will be described in the remainder of this section.

### A. AN OPTIMIZATION PROBLEM
Optimization problems either try to maximize the fitness or minimize the cost. An optimization problem which tries to minimize the cost function can be defined in Eq. (1) as follows:

$$Min\ F(\vec{X}),\ \vec{X} = (X(1), \dots, X(n)),\ X(j) \in [LB(j), UB(j)] \quad (1)$$

where $F(\vec{X})$ is the cost function, $\vec{X}$ is a vector describing the design parameters, and $LB(j)$ and $UB(j)$ are the lower and upper bound of the $j'th$ parameter, respectively.

### B. EXPLANATION OF HARMONY SEARCH (HS) ALGORITHM
Each solution in HS algorithm is called a harmony and described as an n-dimensional vector. First, the initial population of HS should be randomly generated from the search space. Then, a new harmony is constructed by three major operators in HS namely, memory consideration, pitch adjustment, and random re-initialization. The algorithm proceeds by comparing the new harmony ($\vec{X}_{new}$) with the existing worst harmony in the HM ($\vec{X}_{worst}$). $\vec{X}_{new}$ will be substituted by $\vec{X}_{worst}$ provided that the new harmony has better cost than the worst harmony. The process of generating $\vec{X}_{new}$ and possible substitution with $\vec{X}_{worst}$ continues until convergence occurs.

HS has some parameters which should be preallocated at the beginning of optimization. These parameters which tune HS for optimized quality of convergence are harmony memory consideration rate (HMCR), harmony memory size (HMS), number of improvisations (NI), pitch adjustment rate (PAR), and bandwidth (BW).

### C. INITIALIZING HARMONY MEMORY (HM)
Harmony memory (HM) is a matrix in which the rows indicate the harmonies. Suppose that $\vec{X}_i = (X_i(1), \dots, X_i(n))$ is the $i'th$ row (harmony) in HM. Then, each row of HM can be initialized by Eq. (2) as follows:

$$X_i(j) = LB(j) + (UB(j) - LB(j)) \times r,\ for\ i = 1 \dots HMS\ \&\ j = 1 \dots n \quad (2)$$

Where $r$ is a random number in range of [0,1). Hence, HM is a matrix of HMS size with solution vectors in each row as shown in Eq. (3).

$$HM = \begin{bmatrix} X_1(1) & , & X_1(2) & \dots & , X_1(n) \\ X_2(1) & , & X_2(2) & \dots & , X_2(n) \\ & & . & & \\ & & . & & \\ X_{HMS}(1), & X_{HMS}(2), & \dots, & X_{HMS}(n) \end{bmatrix} \quad (3)$$

### D. IMPROVISATION OF A NEW HARMONY
Each element of $\vec{X}_{new}$, ($X_{new}(j)$), is created by Eq. (4) provided that the randomly generated number ($r_1 \in [0,1)$) is less than HMCR. However, assuming $r_1$ is greater than HMCR, Eq. (2) is used for randomly creating $X_{new}(j)$.

$$\begin{cases} X_{new}(j) = X_a(j) \\ a = \lfloor r * HMS + 1 \rfloor \end{cases} \quad (4)$$

Where $r$ is a random number in the range of [0,1). $r_2 \in [0,1)$ is a randomly generated number provided that $X_{new}(j)$ was selected from HM. If $r_2$ is less than pitch adjustment rate (PAR), an adjustment using Eq. (5) must be performed:

$$X_{new}(j) = X_{new}(j) + BW(j) \times r \quad (5)$$

Where $r$ is a random number in the range of [0,1) and $BW(j)$ is the corresponding bandwidth of the $j'th$ parameter in $\vec{X}_{new}$.

### E. UPDATING THE HARMONY MEMORY (HM)
The HM matrix should be updated upon creating a new harmony. HM rows are sorted ascendingly based on the cost. Thus, the last row of HM incorporates the worst harmony ($\vec{X}_{worst}$). The new recently created harmony, $\vec{X}_{new}$, will be substituted by existing $\vec{X}_{worst}$ if $\vec{X}_{new}$ has cheaper cost.in comparison with $\vec{X}_{worst}$. The sorting of HM matrix should be updated accordingly.

## IV. MATERIALS AND METHODS
We have used a harmony search algorithm to minimize error in anemia classification. For this purpose, a new approximation method has been developed as the optimization tool for minimizing the approximator error, based on the harmony search algorithm (HS). The method borrows some principles from a successfully tested structure representation on genetic algorithms for nonlinear function fitting, or STROGANOFF, method [44]. The proposed method is explained in the following subsections.

## A. THE PRINCIPLES OF STROGANOFF

In STROGANOFF, the input-output mapping can be described as a smooth multidimensional surface which plays the role of universal approximator [45]. The method uses a binary tree in which each non-leaf node (which is commonly a second-order polynomial) is modeled by a differentiable function. Additionally, each leaf node contains an input variable of the system. Fig. 1 shows a representation of the system.

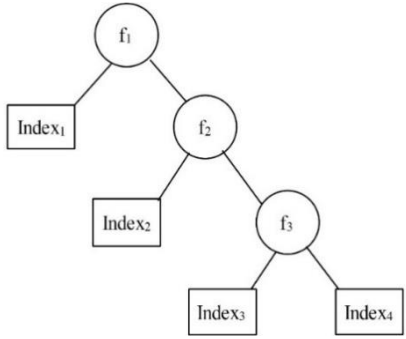

**FIGURE 1.** Binary tree representation of a polynomial with four indices as the input variables.

Suppose that the binary tree (Fig. 1) is used to approximate a function $y = g(x_1, x_2, x_3, x_4) = g(\vec{X})$ where an input-output dataset $\{(\overrightarrow{X_k}, y_k); k = 1..N\}$ is available. The binary tree in Fig. 1 takes four input variables (the elements of $\overrightarrow{X_k} = x_1^k, x_2^k, x_3^k, x_4^k$) on its leaf nodes and produces an approximate output $(\hat{y}_k)$. The produced output can be obtained by Eq. (6) as follows.

$$\hat{y}_k = f_1\left(x_4^k, f_2\left(x_3^k, f_3(x_1^k, x_2^k)\right)\right) \quad (6)$$

Where each $f_i$ takes the bi-variable quadratic polynomial that comes from the Kolmogorov-Gabor polynomial formulated as Eq. (7) [45].

$$f(x) = a_0 + \sum_i a_i x_i + \sum_i \sum_j a_{ij} x_i y_j + \sum_i \sum_j \sum_k a_{ijk} x_i y_j z_k + \cdots \quad (7)$$

Where the $a_i$ are the polynomial coefficients and the $x_i$ are the independent variables. The Kolmogorov-Gabor polynomial is a universal format for function modeling because it can be used to approximate any continuous function mapping to an arbitrary precision, if there are sufficient terms. For two independent variables, Eq. (8) with six terms of the Kolmogorov-Gabor polynomial, has been employed by various researchers[45].

$$f(x) = a_0 h_0(x) + a_1 h_1(x) + \cdots + a_5 h_5(x) \quad (8)$$

Where $a_i \neq 0$, $0 \leq i \leq 5$, and the simple functions $h_i(x)$ that produce the terms are: $h_0(x) = 1$, $h_1(x) = x_1$, $h_2(x) = x_2$, $h_3(x) = x_1 x_2$, $h_4(x) = x_1^2$, and $h_5(x) = x_2^2$. Hence, the polynomial can be written as Eq. (9).

$$f(x) = a_0 + a_1 x_1 + a_2 x_2 + a_3 x_1 x_2 + a_4 x_1^2 + a_5 x_2^2 \quad (9)$$

For the current optimization problem, the coefficients of the polynomials in non-leaf nodes of the binary tree (Fig. 1) must be tuned up by fitting the output of function $(\hat{y}_k)$ to the desired output $(y_k)$. Nikolay Y. Nikolaev and Hitoshi Iba [45] have proposed 16 quadratic forms presented in Table 1 to be applied on the non-leaf nodes of STROGANOFF algorithm instead of the complete bi-variable quadratic polynomial (Eq. 9).

TABLE 1.
BIVARIATE BASIS POLYNOMIALS USED IN STROGANOFF.

| index | Polynomial function | Number of terms |
|---|---|---|
| 1 | $f_1(X^*) = a_0 + a_1 x_1 + a_2 x_2 + a_3 x_1 x_2$ | 4 |
| 2 | $f_2(X) = a_0 + a_1 x_1 + a_2 x_2$ | 3 |
| 3 | $f_3(X) = a_0 + a_1 x_1 + a_2 x_2 + a_3 x_1^2 + a_4 x_2^2$ | 5 |
| 4 | $f_4(X) = a_0 + a_1 x_1 + a_2 x_1 x_2 + a_3 x_1^2$ | 4 |
| 5 | $f_5(X) = a_0 + a_1 x_1 + a_2 x_2^2$ | 3 |
| 6 | $f_6(X) = a_0 + a_1 x_1 + a_2 x_2 + a_3 x_1^2$ | 4 |
| 7 | $f_7(X) = a_0 + a_1 x_1 + a_2 x_1^2 + a_3 x_2^2$ | 4 |
| 8 | $f_8(X) = a_0 + a_1 x_1^2 + a_2 x_2^2$ | 3 |
| 9 | $f_9(X) = a_0 + a_1 x_1 + a_2 x_2 + a_3 x_1 x_2 + a_4 x_1^2 + a_5 x_2^2$ | 6 |
| 10 | $f_{10}(X) = a_0 + a_1 x_1 + a_2 x_2 + a_3 x_1 x_2 + a_4 x_1^2$ | 5 |
| 11 | $f_{11}(X) = a_0 + a_1 x_1 + a_2 x_1 x_2 + a_3 x_1^2 + a_4 x_2^2$ | 5 |
| 12 | $f_{12}(X) = a_0 + a_1 x_1 x_2 + a_2 x_1^2 + a_3 x_2^2$ | 4 |
| 13 | $f_{13}(X) = a_0 + a_1 x_1 + a_2 x_1 x_2 + a_3 x_2^2$ | 4 |
| 14 | $f_{14}(X) = a_0 + a_1 x_1 + a_2 x_1 x_2$ | 3 |
| 15 | $f_{15}(X) = a_0 + a_1 x_1 x_2$ | 2 |
| 16 | $f_{16}(X) = a_0 + a_1 x_1 x_2 + a_2 x_1^2$ | 3 |

\* $X = (x_1, x_2)$

## B. CHOOSING THE INDICES OF CBC AS THE SYSTEM'S INPUTS

After reviewing CBC indices, the four indices Hb, RBC, MCV, and HCT are chosen as the system's inputs. Indices are chosen for minimum redundancy. To achieve this goal, we use the algorithm described in the two next subsections and some sample collected CBC specimens, drawn from those shown in Table 2. The chosen indices can populate the leaf nodes of a binary tree such as the one shown in Fig. 1. Additionally, these indices have been successfully used in various previous studies [2], [46].

### 1) DATA ANALYSIS

The dataset used in this work has been obtained from [47]. The total number of collected CBC tests is 750, obtained from the blood specimens of 390 males aged 20-35 and 360 females aged 17-32 years. The information is organized in three classes as follows [47].

I. The samples with CBC indices in the normal ranges (218 samples).
II. The samples with significantly abnormal CBC indices, identified definitively as β-TT or IDA (98 samples, IDA = 38, β-TT = 60).
III. The samples with indeterminate CBC indices. They have been determined by complementary tests such as electrophoresis (434 samples, IDA = 231, β-TT = 203).

Based on the data, the total numbers of β-TT and IDA subjects are 269 and 263, respectively. Table 2 shows some examples of different classes of collected samples. Abnormal values are shown in bold face (Table 2). It is clear that the samples in class III are sufficiently similar and the relationship is sufficiently complex that a simple mathematical formula cannot distinguish reliably between IDA and β-TT subjects in this class. As the study's purpose is to differentiate between IDA and β-TT subjects, we excluded the normal samples and the remaining samples (532 samples including IDA and β-TT) were randomly divided into the training and the testing sets of 132 and 400 samples, respectively.

### 2) AN ALGORITHM FOR ELIMINATION REDUNDANT INDICES

The elimination of redundant CBC indices from the input set leads to a more efficient system. Fig. 2 shows the algorithm which is employed for choose the most reliable indices for discrimination. this algorithm is called Pattern Based Index Selection (PBIS). The name is chosen because the elimination is conducted based on the similarity between indices with respect to their abnormal patterns, as seen in the Table 2.

As seen in Fig. 2, for each iteration of the algorithm, the index such has the least similarity with the other indices, with respect to their abnormality pattern, must be found. Eq. (10) can be used to calculate the Coefficient of Similarity (COS) between any two CBC indices.

$$COS_{i,j} = 2 - \frac{\sum_{k=1}^{n}\left(\left|Lo_i(x_i^k) - Lo_j(x_j^k)\right| + \left|Hi_i(x_i^k) - Hi_j(x_j^k)\right|\right)}{n} \quad (10)$$

Where $n$ is the number of samples, $x_i^k$ is the value of the $i'th$ CBC index for the $k'th$ sample, and $Lo_i(x)$ and $Hi_i(x)$ are functions that denote whether $x$ is below the lower bound or above the upper bound of the normal range for the $i'th$ CBC index, respectively. $Lo_i(x)$ and $Hi_i(x)$ can be obtained from Eqs. (11,12) as follows.

$$Lo_i(x) = \begin{cases} 1 & if\ x < lower\ bound\ of\ normal\ range \\ & for\ i'th\ CBC\ index \\ 0 & if\ x \geq lower\ bound\ of\ normal\ range \\ & for\ i'th\ CBC\ index \end{cases} \quad (11)$$

$$Hi_i(x) = \begin{cases} 1 & if\ x > upper\ bound\ of\ normal\ range \\ & for\ i'th\ CBC\ index \\ 0 & if\ x \leq upper\ bound\ of\ normal\ range \\ & for\ i'th\ CBC\ index \end{cases} \quad (12)$$

After applying the above algorithm, four indices, RBC, Hb, HCT, and MCV, are chosen as system inputs.

TABLE 2.
SEVERAL EXAMPLES OF COLLECTED CBC INDICES [47].

| RBC(M/mm3) | Hb(g/dL) | HCT (%) | MCV(fL) | MCH(pg) | MCHC (%) | Class | Type |
|---|---|---|---|---|---|---|---|
| (4.5-6.3) * | (13.5-18) | (39-50) | (80-96) | (27-32) | (32-38) | | |
| 5.43 | 10.2 ↓ | 34 ↓ | 62.6 ↓ | 18.8 ↓ | 30 ↓ | II | βTT |
| 6.13 | 12.5 ↓ | 40.1 | 65.4 ↓ | 20.4 ↓ | 31.2 ↓ | II | βTT |
| 6.8 ↑ | 12.6 ↓ | 43.4 | 63.8 ↓ | 18.5 ↓ | 29 ↓ | II | βTT |
| 4.73 | 10.1 ↓ | 38.7 ↓ | 69.9 ↓ | 19.7 ↓ | 30.3 ↓ | II | IDA |
| 4.13 ↓ | 8.2 ↓ | 38.4 ↓ | 71.8 ↓ | 19.9 ↓ | 28.9 ↓ | II | IDA |
| 4.61 | 12.8 ↓ | 38.9 ↓ | 84.4 | 27.8 | 32.9 | I | nor |
| 4.36 ↓ | 13.1 ↓ | 39.7 | 91.1 | 30 | 33 | I | nor |
| 4.77 | 13.3 ↓ | 39.7 | 83.2 | 27.9 | 33.5 | I | nor |
| 3.99 ↓ | 11.4 ↓ | 35.1 ↓ | 88 | 28.6 | 32.5 | I | nor |
| 4.4 ↓ | 13.7 | 40.5 | 92 | 31.1 | 33.8 | I | nor |
| 6.62 ↑ | 13.1 ↓ | 43.9 | 65.9 ↓ | 19.7 ↓ | 29.8 ↓ | III | βTT |
| 4.56 | 10.3 ↓ | 34.7 ↓ | 76.1 ↓ | 22.6 ↓ | 29.7 ↓ | III | βTT |
| 5.95 | 12.5 ↓ | 43.4 | 72.8 ↓ | 21 ↓ | 28.9 ↓ | III | βTT |
| 4.65 | 10.5 ↓ | 35.6 ↓ | 76.6 ↓ | 22.6 ↓ | 29.5 ↓ | III | βTT |
| 5.12 | 10.6 ↓ | 36.7 ↓ | 71.7 ↓ | 20.7 ↓ | 28.9 ↓ | III | βTT |
| 4.81 | 11.6 ↓ | 37.6 ↓ | 78.2 ↓ | 24.1 ↓ | 30.9 ↓ | III | IDA |
| 5.05 | 11.5 ↓ | 38.8 ↓ | 76.8 ↓ | 22.8 ↓ | 29.6 ↓ | III | IDA |
| 4.88 | 12.1 ↓ | 38.8 ↓ | 79.5 ↓ | 24.8 ↓ | 31.2 ↓ | III | IDA |
| 5.03 | 12.7 ↓ | 39.9 | 79.3 ↓ | 25.2 ↓ | 31.8 ↓ | III | IDA |
| 5.1 | 13.2 ↓ | 39.7 | 77.8 ↓ | 25.9 ↓ | 30.7 ↓ | III | IDA |

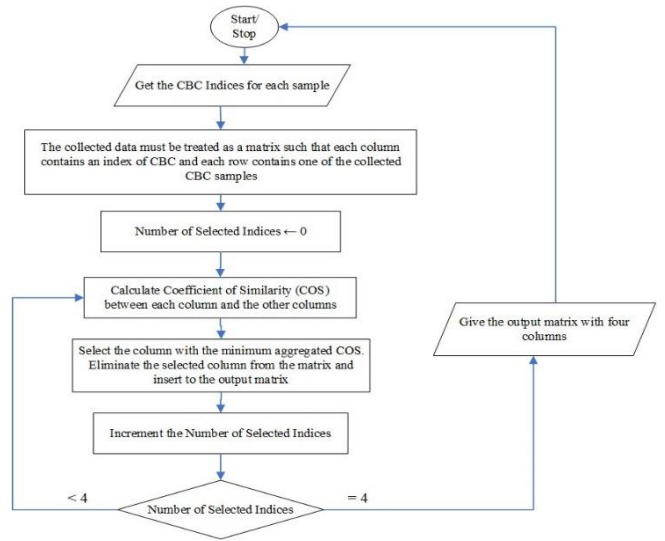

FIGURE 2. The algorithm for choosing inputs

## C. PROPOSED METHOD

The proposed method is based on harmony search and certain genetic programming principles. Before defining a harmony, the elements of the binary tree (Fig. 1) must be defined. The four indices of CBC chosen in the previous section populate the leaf nodes. Non-leaf nodes take the complete bi-variable polynomial that comes from the Kolmogorov-Gabor polynomial with six coefficients (Eq. 9).

There are three non-leaf nodes in the tree and the polynomial shown in (Eq. 9) is applied to each of them. Hence, there are 18 parameters that must be tuned up to reach the optimum value. Because $f_i$ is a symmetric function, we can generate 15 different schemes for the tree (Fig. 1) and the four CBC indices chosen as input variables. Table 3 lists the possible schemes for the tree.

TABLE 3.
DIFFERENT SCHEMES FOR GENERATING THE TREE WITH CHOSEN CBC.

| Id | Scheme | Function |
|---|---|---|
| 0 |  | $f_1(f_2(RBC,Hb),f_3(HCT,MCV))$ |
| 1 |  | $f_1(f_2(RBC,HCT),f_3(Hb,MCV))$ |
| 2 |  | $f_1(f_2(RBC,MCV),f_3(Hb,HCT))$ |
| 3 |  | $f_1(f_2(f_3(RBC,Hb),HCT),MCV)$ |
| 4 |  | $f_1(f_2(f_3(RBC,Hb),MCV),HCT)$ |
| 5 |  | $f_1(f_2(f_3(RBC,HCT),Hb),MCV)$ |
| 6 |  | $f_1(f_2(f_3(RBC,HCT),MCV),Hb)$ |
| 7 |  | $f_1(f_2(f_3(RBC,MCV),Hb),HCT)$ |
| 8 |  | $f_1(f_2(f_3(RBC,MCV),HCT),Hb)$ |
| 9 |  | $f_1(f_2(f_3(Hb,HCT),RBC),MCV)$ |
| 10 |  | $f_1(f_2(f_3(Hb,HCT),MCV),RBC)$ |
| 11 |  | $f_1(f_2(f_3(Hb,MCV),RBC),HCT)$ |
| 12 |  | $f_1(f_2(f_3(Hb,MCV),HCT),RBC)$ |
| 13 |  | $f_1(f_2(f_3(HCT,MCV),RBC),Hb)$ |
| 14 |  | $f_1(f_2(f_3(HCT,MCV),Hb),RBC)$ |

A one-dimensional array is assigned to the harmony, containing 19 elements. Fig. 3 shows the fields of a harmony for the proposed method.

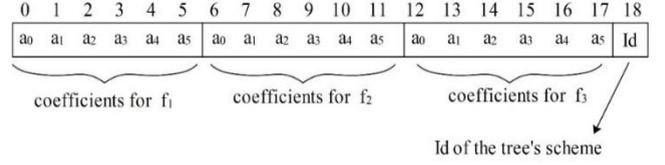

FIGURE 3. Structure of a harmony for the proposed method

Therefore, a harmony describes the structure of a tree, which can be one of the 15 different schemes in Table 3, and the $f_i$ coefficients of that tree. To determine the most suitable scheme for this problem, the harmony search algorithm can use a heuristic search to find the best structure for the system tree using the 'Id' field of the harmony and evaluating its corresponding tree to optimize the coefficients of the polynomials of the tree.

## D. IMPLEMENTING THE PROPOSED METHOD

We use the programming language C# to implement the proposed method. A user interface is provided for quick testing and tuning the algorithm's parameters. The interface gives the user the ability to set various parameters, including the size of harmony memory (HMS), the initial values for HMCR and PAR, the number of improvisations (NI), and the choice of standard harmony search or dynamic harmony search (DHS). Additionally, the interface monitors the outputs of the program.

### 1) INITIALIZE THE HARMONY MEMORY

The harmony memory is initialized randomly by Eq. (2). The default value for HMS is 100, but it can be changed by the user. Hence, the harmony memory is a matrix of floating-point numbers of size 100×20. The elements of column 20 of the HM contain the fitness values for of each row/harmony. As shown in Fig. 3, a harmony is a one-dimensional array of size 19 where the first 18 elements contain the polynomials' coefficients and the last element contains the Id of the tree scheme. Hence, the values for the 18 first elements can be determined by trial and error, but the last element must be an integer in range of [0, 14].

Therefore, we can calculate the values for the first 18 elements of harmony by Eq. (2), but the last element is obtained by Eq. (13) as follows:

$$X_i(j) = ceil(15 \times r), \quad for\ i = 1..HMS\ , j = 19$$
(13)

Where $r$ is a random number in range of [0, 1]. The values for column 20 of the HM are determined by a fitness function described in the next section. Next, the HM rows are sorted in descending order with respect to their fitness (column 20). The harmonies in the HM are now organized best to worst from top to bottom.

## 2) CALCULATING THE FITNESS FOR EACH HARMONY

To determine the best harmony for the discrimination of IDA from thalassemia trait, suppose that $\alpha$ and $\beta$ are the averages for the tree's outcome when the CBC indices of the samples with IDA and thalassemia trait are applied, and $\gamma$ and $\delta$ are the variances of those values, respectively. It is clear that a function with a larger distance between averages and smaller variances is a better discriminator. Therefore, we use Eq. (14) as the fitness function for the harmony.

$$fitness = \frac{|\alpha - \beta|}{1 + \gamma \delta} \qquad (14)$$

## 3) USING DYNAMIC HARMONY SEARCH (DHS) INSTEAD OF STANDARD HS

Dynamic harmony search combined with tuning the parameters leads to faster convergence [48]. In the standard HS, the parameters of the algorithm such as HMCR and PAR have fixed values, but in the DHS, the parameter HMCR changes with Eq. (15) as follows:

$$HMCR_{new} = HMCR_{old} + (1 - HMCR_{old}) \times (\alpha - \beta) \qquad (15)$$

where $\alpha$ and $\beta$ are the normalized fitness/cost average of ten recently considered memory solutions and ten recently generated random solutions respectively in maximum finding. However, In case of minimum finding, the definitions of solutions are contrariwise. This improvement in the HS algorithm is tunned to be occurred every ten iterations. Likewise, variable $E$ (which is derived from memory consideration) is used to evaluate the adjustment of new solutions for improving the pitch adjustment rate (PAR) as stated in Fig. 4. Variable $E$ can be further used in Eq. (16) to improve the parameter PAR for the proposed harmony search algorithm.

$$PAR_{new} = PAR_{old} + \frac{E \times (1 - PAR_{old})}{100} \qquad (16)$$

Fig. 5 illustrates the improving process of PAR for every 10 recently generated memory solutions. For comparison between standard HS and DHS, the user interface can perform the proposed method with both algorithms optionally. As can be seen in Fig. 4, the standard HS and DHS both have converged to the optimal value, but the standard HS needs more time than the dynamic harmony search.

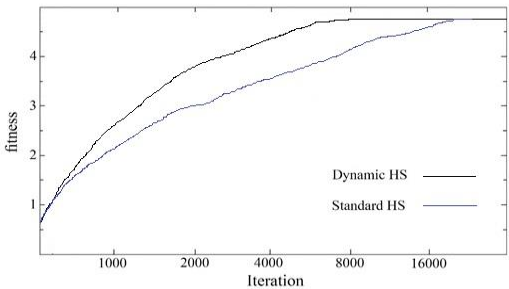

**FIGURE 4.** Comparison between the standard HS and Dynamic HS (DHS)

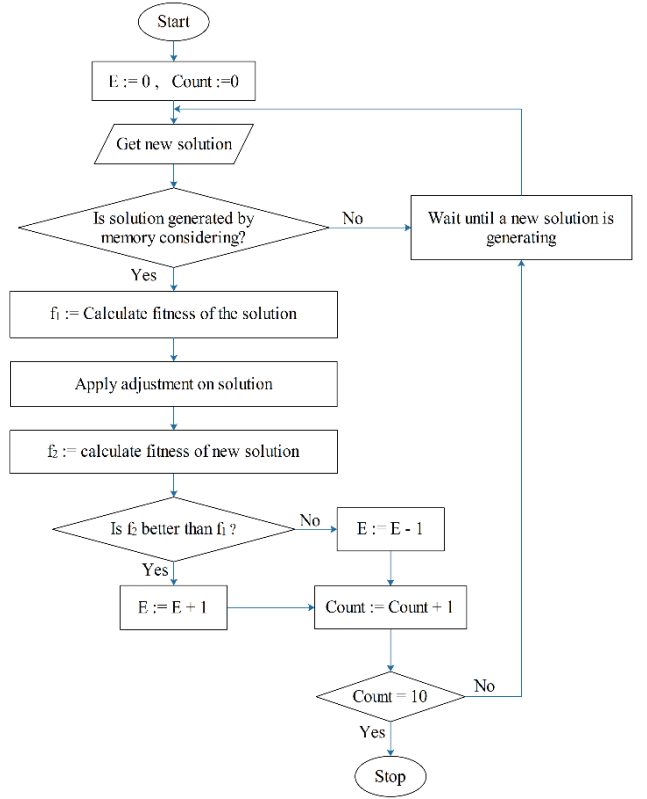

**FIGURE 5.** Magnetization as a function of applied field. Note that "Fig." is abbreviated. There is a period after the figure number, followed by two spaces. It is good practice to explain the significance of the figure in the caption.

### E. TESTING THE PROPOSED METHOD

To test the proposed method, 750 CBC samples have been obtained from [47]. After eliminating unwanted samples, 532 CBC samples remain for further examination. Of the remaining subjects, there are 269 IDA and 263 β-TT subjects. Six popular medical metrics, sensitivity (SENS), specificity (SPEC), positive predictive value (PPV), negative predictive value (NPV), accuracy (ACC) and Youden's index (YI), have been used in all of the experiments. These metrics are obtained by Eqs. (17-22).

$$SENS = \frac{TP}{TP+FN} \times 100 \qquad (17)$$

$$SPEC = \frac{TN}{TN+FP} \times 100 \qquad (18)$$

$$PPV = \frac{TP}{TP+FP} \times 100 \qquad (19)$$

$$NPV = \frac{TN}{TN+FN} \times 100 \qquad (20)$$

$$ACC = \frac{TP+TN}{TP+TN+FP+FN} \times 100 \qquad (21)$$

$$YI = SENS + SPEC - 100 \qquad (22)$$

Where TP, TN, FP and FN are true positive, true negative, false positive and false negative, respectively. Table 4 shows the results of applying the method to the test samples.

TABLE 4.
COMPARISON OF THE METHOD'S IDA DIAGNOSIS RESULTS WITH DIFFERENT PARAMETERS (NUMBER OF SAMPLES = 400, IDA = 207, B-TT = 193).

| Item | NI* | HMS** | TP | TN | FP | FN | SENS (%) | SPEC (%) | PPV (%) | NPV (%) | ACC (%) | YI (%) |
|---|---|---|---|---|---|---|---|---|---|---|---|---|
| Proposed method | 2000 | 50 | 191 | 184 | 9 | 16 | 92.3 | 95.3 | 95.5 | 92.0 | 93.7 | 87.6 |
| | 2000 | 100 | 194 | 186 | 7 | 13 | 93.7 | 96.4 | 96.5 | 93.5 | 95.0 | 90.1 |
| | 8000 | 50 | 198 | 189 | 4 | 9 | 95.6 | 97.9 | 98.0 | 95.5 | 96.7 | 93.6 |
| | 8000 | 100 | 201 | 191 | 2 | 6 | 97.1 | 98.9 | 99.0 | 96.9 | 98.0 | 96.1 |
| STROGANOFF | 2000 | 50 | 185 | 181 | 12 | 22 | 89.4 | 93.4 | 93.9 | 89.2 | 91.5 | 83.2 |
| | 2000 | 100 | 192 | 187 | 6 | 15 | 92.8 | 96.9 | 97.0 | 92.6 | 94.7 | 89.7 |
| | 8000 | 50 | 189 | 187 | 6 | 18 | 91.3 | 96.9 | 96.9 | 91.2 | 94.0 | 88.2 |
| | 8000 | 100 | 195 | 189 | 4 | 12 | 94.2 | 97.9 | 98.0 | 94.0 | 96.0 | 92.1 |

\* for STROGANOFF, it is number of generations
\*\* for STROGANOFF, it is population size

## V. COMPARISON

This section compares the best results of the proposed method against well-known works in the literature. This section is organized in two subsections. The first compares the proposed method with some traditional and some more recent methods on the same dataset. The second subsection compares the performance of the proposed method with the information reported by other researchers.

### A. Traditional method for discrimination between IDA and β-TT

There are several mathematical methods that traditionally have been used for the IDA/ β-TT discrimination problem. Some of these methods are shown in Table 5. For better comparison, the traditional methods were tested on the same CBC samples used for the proposed method. The results for these methods are shown in Table 6. The results in Table 6 show that the proposed method is significantly more accurate than traditional methods. The range of accuracy for the traditional methods, which are based on a simple mathematical formula, is in (78, 93.2), while the proposed method has an accuracy of approximately 98%.

TABLE 5.
TRADITIONAL METHOD FOR DISCRIMINATION BETWEEN IDA AND B-TT

| Authors | Symbol | Year | Formula | IDA | β-TT |
|---|---|---|---|---|---|
| Mentzer et al. [14] | MI | 1973 | $MCV/RBC$ | $\geq 13$ | $< 13$ |
| England & Fraser [7] | E&FI | 1973 | $MCV - RBC - 5 \times Hb - 3.4$ | $\geq 0$ | $< 0$ |
| Srivastava & Bevington [49] | S&BI | 1973 | $MCH/RBC$ | $\geq 3.8$ | $< 3.8$ |
| Shine & Lal [12] | S&LI | 1977 | $MCV^2 \times MCH \times 0.01$ | $\geq 1530$ | $< 1530$ |
| Sirdah et al. [16] | SI | 2008 | $MCV - RBC - 3 \times Hb$ | $\geq 27$ | $< 27$ |
| Ehsani et al. [6] | EI | 2009 | $MCV - 10 \times RBC$ | $\geq 15$ | $< 15$ |
| Green & King [8] | G&KI | 1989 | $MCV^2 \times RDW \times Hb \times 0.01$ | $\geq 72$ | $< 72$ |

TABLE 6.
COMPARISON OF TRADITIONAL METHODS AND THE PROPOSED METHOD IN DISCRIMINATION BETWEEN IDA (SAMPLES ARE SAME AS TABLE 4).

| Method | TP | TN | FP | FN | SENS (%) | SPEC (%) | PPV (%) | NPV (%) | ACC (%) | YI (%) |
|---|---|---|---|---|---|---|---|---|---|---|
| MI [14] | 175 | 165 | 28 | 32 | 84.5 | 85.5 | 86.2 | 83.8 | 85.0 | 70.0 |
| E&FI [7] | 152 | 161 | 32 | 55 | 73.4 | 83.4 | 82.6 | 74.5 | 78.0 | 56.8 |
| S&BI [49] | 161 | 161 | 32 | 46 | 77.8 | 83.4 | 83.4 | 77.8 | 80.5 | 61.2 |
| S&LI [12] | 190 | 183 | 10 | 17 | 91.8 | 94.8 | 95.0 | 91.5 | 93.2 | 86.6 |
| SI [16] | 155 | 157 | 36 | 52 | 74.9 | 81.3 | 81.2 | 75.1 | 78.0 | 56.2 |
| EI [6] | 160 | 168 | 25 | 47 | 77.3 | 87.0 | 86.5 | 78.1 | 82.0 | 64.3 |
| G&KI [8] | 184 | 173 | 20 | 23 | 88.9 | 89.6 | 90.2 | 88.3 | 89.3 | 78.5 |
| This work | 201 | 191 | 2 | 6 | 97.1 | 98.9 | 99.0 | 96.9 | 98.0 | 96.1 |

### B. Artificial intelligence-based models for classification anemia

There are several AI-based models for the diagnosis and classification of anemia. The second set of experiments compares the performance of the proposed method with other, similar methods. As can be observed in Table 7, the proposed method is more accurate than these other similar methods. Although the results for [50] seem to be better than this work, but the method which is used by [50] get thee image of blood cells as the system input instead of CBC, which needs additional laboratory test on patient.

TABLE 7.
COMPARISON OF THIS WORK AND REPORTED RESULTS OF SOME SIMILAR METHODS.

| Authors | Year | Method | Sensitivity | Specificity | Accuracy |
|---|---|---|---|---|---|
| Azarkhish et al. [33] | 2012 | ANFIS[1] | 87.1 | 95.6 | 90.7 |
| Azarkhish et al. [33] | 2012 | ANN[2] | 96.8 | 95.6 | 96.3 |
| Amendolia et al. [32] | 2002 | MLP[3] | 95.0 | 92.0 | 93.5 |
| Amendolia et al. [31] | 2003 | SVM[4] | 95.0 | 83.0 | 89.0 |
| Amendolia et al. [31] | 2003 | KNN[5] | 77.0 | 93.0 | 85.0 |
| Khaki Jamei et al. [47] | 2016 | PBIS-ANN[6] | 97.7 | 98.4 | 98.0 |
| Masala et al. [3] | 2013 | RBF[7] | 93.0 | 91.0 | ----- |
| Masala et al. [3] | 2013 | PNN[8] | 89.0 | 73.0 | ----- |
| Shikha Purwar et al. [50] | 2019 | KNN[5], SVM[4], ANN[2] | 98.0 | 99.0 | 99.0 |
| This Work | | DHS[9] | 97.1 | 98.9 | 98.7 |

[1]adaptive neuro-fuzzy inference system; [2]artificial neural network; [3]multi-layer perceptron; [4]support vector machine; [5]K-nearest neighbor; [6]pattern-base input selection artificial neural network; [7]radial basis function; [8]probabilistic neural network; [9]dynamic harmony search ;

## VI. CONCLUSIONS

In this study, a new method based on dynamic harmony search (DHS) was proposed for the discrimination of iron deficiency anemia (IDA) and β-thalassemia trait (β-TT). The method is implemented in C# and has been successfully tested using collected CBC sample data as input. Choosing the most suitable CBC indices to use as the system input is performed by a pattern-based index selection (PBIS) algorithm. The proposed method has been trained on 132 CBC samples. The results indicate that the proposed method, with an accuracy of approximately 98%, outperforms the other methods in the literature. The existing artificial neural network (ANN), MLP, and ANFIS methods, respectively, show the nearest performance on the anemia classification problem.